\documentclass[3p,times,procedia]{elsarticle}
\usepackage{ecrc}
\usepackage[bookmarks=false]{hyperref}
    \hypersetup{colorlinks,
      linkcolor=blue,
      citecolor=blue,
      urlcolor=blue}

\volume{00}

\firstpage{1}

\journalname{Procedia Computer Science}

\runauth{Yuqicheng Zhu et al.}

\jid{procs}

\usepackage{amssymb}
\usepackage{algorithm}
\usepackage{algpseudocode}
\usepackage{graphicx}
\usepackage{textcomp}
\usepackage{xcolor}
\usepackage{amsmath,amssymb,amsfonts}
\usepackage{multicol}

\usepackage[figuresright]{rotating}

\begin{document}
\begin{frontmatter}



\dochead{4th International Conference on Industry 4.0 and Smart Manufacturing}

\title{Active Transfer Prototypical Network: An Efficient Labeling Algorithm for Time-Series Data}


\author[a,b]{Yuqicheng Zhu$^{*}$}
\author[a,c]{Mohamed-Ali Tnani}
\author[b]{Timo Jahnz}
\author[a]{Klaus Diepold}

\address[a]{Department of Electrical and Computer Engineering, Technical University of Munich, Arcisstraße 21, 80333 Munich, Germany}
\address[b]{Department of Brake System Engineering, Robert Bosch GmbH, Robert-Bosch-Allee 1, 74232 Abstatt, Germany}
\address[c]{Department of Factory of the Future, Bosch Rexroth AG, Lise-Meitner-Straße 4, 89081 Ulm, Germany}

\correspondingauthor[*]{ Corresponding author. Tel.: +49 711 685 88127. {\it E-mail address:} yuqicheng.zhu@ipvs.uni-stuttgart.de}

\begin{abstract}
The paucity of labeled data is a typical challenge in the automotive industry. Annotating time-series measurements requires solid domain knowledge and in-depth exploratory data analysis, which implies a high labeling effort. Conventional \textit{Active Learning} (AL) addresses this issue by actively querying the most informative instances based on the estimated classification probability and retraining the model iteratively. However, the learning efficiency strongly relies on the initial model, resulting in the trade-off between the size of the initial dataset and the query number. This paper proposes a novel \textit{Few-Shot Learning} (FSL)-based AL framework, which addresses the trade-off problem by incorporating a \textit{Prototypical Network} (ProtoNet) in the AL iterations. The results show an improvement, on the one hand, in the robustness to the initial model and, on the other hand, in the learning efficiency of the ProtoNet through the active selection of the support set in each iteration. This framework was validated on UCI HAR/HAPT dataset and a real-world braking maneuver dataset. The learning performance significantly surpasses traditional AL algorithms on both datasets, achieving 90\% classification accuracy with 10\% and 5\% labeling effort, respectively.
\end{abstract}


\end{frontmatter}


\section{Introduction}
The rapid growth of digitalization and information systems allows industries to collect vast amounts of data from devices such as machines and sensors. In particular, in the automotive industry, modern vehicles generate gigabytes of time-series data every hour. This large amount of data drives the development of innovative applications that support the user, accelerate innovation cycles, and enhance the entire workflow in manufacturing. 

Although time-series classification problem has been well studied for non deep learning classifiers \cite{middlehurstHIVECOTENewMeta2021, bagnallTimeSeriesClassificationCOTE2015, hillsClassificationTimeSeries2014} and deep learning classifiers \cite{fawazDeepLearningTime2019}, a sufficient amount of labeled training data is required for both methods to achieve a satisfying performance. Enormous unlabeled data are frequently available while obtaining a complete set of labeled data might be complicated or expensive. In the automotive industry, especially for the braking system, the amount of sensor signals collected daily by test vehicles is increasing enormously. Due to the complex interconnection of the sensor signals under various braking scenarios, systematic signal analysis and detailed discussion between system experts and application engineers are frequently required to determine the corresponding braking maneuvers. The solid domain knowledge and the high time required for time-series labeling have made the large-scale application of state-of-the-art machine learning techniques in the automotive industry challenging.

Labeling effort of the time-series data can be reduced through \textit{Active Learning} (AL) \cite{angluinQueriesConceptLearning1988} or \textit{Few-Shot Learning} (FSL) \cite{wangGeneralizingFewExamples2020}. AL addresses the labeling issue by actively querying the most informative instances to be labeled from the unlabeled data pool \cite{settlesActiveLearningLiterature}. As the number of manually annotated instances increases, the classification accuracy is expected to improve rapidly. However, the learning performance heavily relies on the initial model. A poor initial model leads to querying several uninformative instances at the beginning \cite{yuanInitialTrainingData2011}. Therefore, high learning efficiency can not be guaranteed due to the random nature of the initial model. On the other hand, FSL attempts to generalize the prior knowledge quickly to new tasks using a few labeled instances, i.e., a support set \cite{wangGeneralizingFewExamples2020}. It has a pre-trained encoder trained on a large labeled dataset of similar tasks, which is used to convert the input instances into embedding space. A new test sample is classified based on the similarity to the embedding of the support set. Nevertheless, an unrepresentative support set may lead to inaccurate estimates of similarity and therefore fail to generalize prior knowledge.

\begin{figure}[ht]
\includegraphics[width=.6\textwidth]{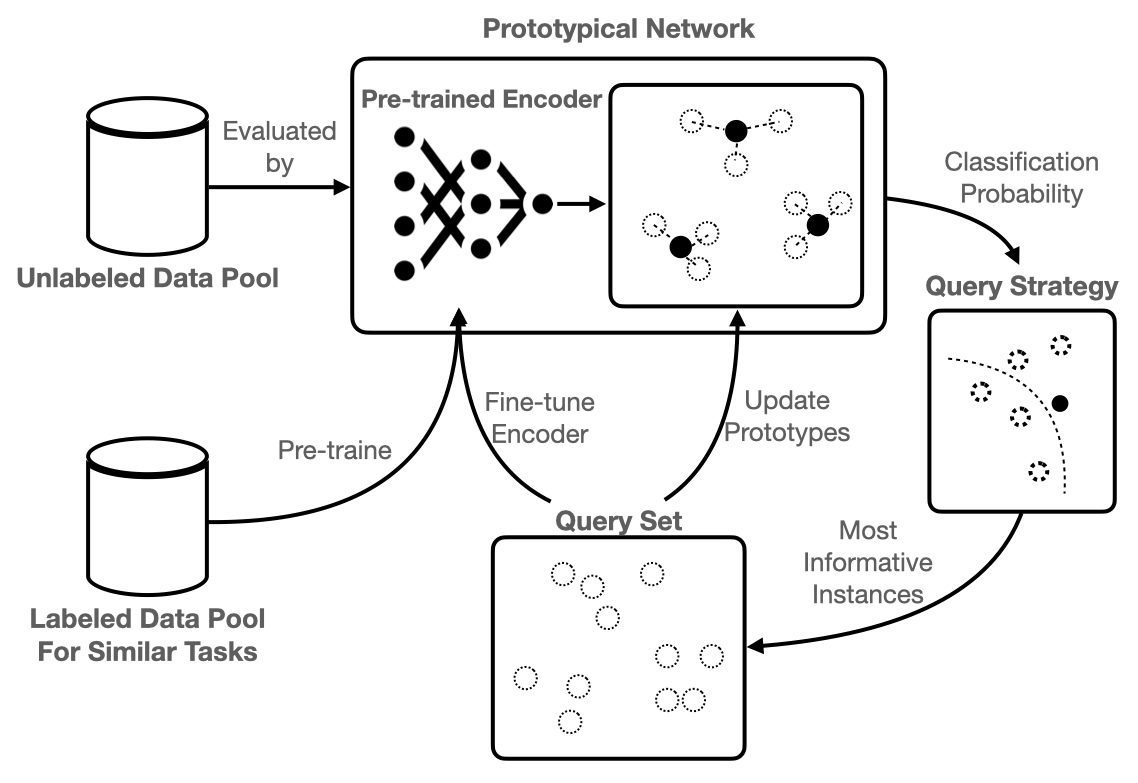}
\centering
\caption{Overview of the algorithm architecture.}
\label{fig:overview}
\end{figure}

The combination of AL and FSL incorporates the advantages and compensates for the disadvantages from both sides. This paper proposes a novel AL framework based on an FSL algorithm, i.e., \textit{Prototypical Network} (ProtoNet) \cite{snellPrototypicalNetworksFewshot2017}. Fig. \ref{fig:overview} visualizes the algorithm architecture. It starts from an unlabeled data pool and a ProtoNet with a pre-trained encoder as the initial model. The goal is to annotate all instances in the unlabeled data pool with minimal manual effort. The ProtoNet evaluates the entire unlabeled data pool. Then, the query strategy picks up the most informative instances based on the estimated classification probabilities. These instances are labeled by experts and used in two parts of the ProtoNet - recalculating the prototypes and fine-tuning the encode. After updating, the next AL iteration starts. The algorithm evaluates the rest of the unlabeled instances through the updated ProtoNet, picks up the most informative instance, and updates the ProtoNet with the accumulated query set. The iteration repeats until it achieves the maximal query number. 

We implemented three different variants of the AL-ProtoNet combination to compare with standard AL and passive learning on public time-series datasets - \textit{UCI HAR\&HAPT} dataset. The influence of different pre-trained encoders and query strategies is investigated. Furthermore, the algorithm with the optimal hyperparameters was validated on a real-world Bosch braking maneuver dataset.

The rest of the paper is organized as follows. Section \ref{relatedwork} sorts out the related works on AL, FSL, and their combination for time-series classification. Relevant background is provided in section \ref{background}. We introduce our proposed novel AL algorithm and other two variants in detail in section \ref{methodology}. The investigation focuses on the learning performance of algorithm structure, query strategy, and pre-trained encoder. Finally, section \ref{exp} describes the experimental settings, followed by the corresponding results and conclusions of our work.

\section{Related Work}\label{relatedwork}
The concept of AL was introduced in 1988 by Angluin, where several query approaches for the problem of using queries to learn an unknown concept have been investigated. \cite{angluinQueriesConceptLearning1988} The objective of AL is to reduce the training set size by involving human annotators in the AL loop. State-of-the-art AL algorithms have been combined with other advanced machine learning approaches to adapt the properties of specific tasks. For example, Wei et al. used semi-supervised learning to recognize unlabeled instances automatically \cite{Wei2006}; Bengar et al. and Emam et al. integrated self-supervised learning in the AL framework \cite{bengar2021, emam2021active}. However, the performance of the algorithm is strongly dependent on a randomly selected initial set. Therefore, our work addresses this problem by leveraging the prior information to select a representative initial data set.

FSL tackles the issue of learning new concepts efficiently by exploiting an encoder trained on a variety of tasks \cite{wangGeneralizingFewExamples2020}. With this pre-trained encoder as the initial model, a classifier can be adapted to previously unseen classes from just a few instances \cite{brenden2015fewshot}. Concretely, \textit{Siamese Network} contains two or more identical sub-networks to find the similarity of the inputs by comparing its feature vectors \cite{Koch2015SiameseNN}. However, using pointwise learning, \textit{Siamese Network} produces a non-probabilistic output and slows the classification task significantly as it requires quadratic pairs to learn. A ProtoNet was proposed by Snell et al., where a metric (embedding) space is learned by computing distances to prototype representations of each class \cite{snellPrototypicalNetworksFewshot2017}. It has been well studied mainly in computer vision to deal with problems such as character recognition \cite{souibgui2021,asish2020}, image classification \cite{snellPrototypicalNetworksFewshot2017,hou2021}, object recognition \cite{prabhudesai2020}, etc. 

Nevertheless, insufficient research has been conducted for time-series classification, even though it is one of the most relevant tasks in the industry. Zhang et al. proposed a novel attentional ProtoNet - \textit{TapNet} for multivariant time-series classification and extended it into a semi-supervised setting \cite{Zhang_Gao_Lin_Lu_2020}. The \textit{Semi-TapNet}, which utilizes the unlabeled data to improve the classification performance, inspires the idea of combining ProtoNet with AL. The idea of active ProtoNet has been investigated and demonstrated to outperform the baseline in \textit{Computer Vision} \cite{WoodwardF17} and \textit{Natural Language Processing} \cite{muller2022active} field. However, Pezeshkpour et al. pointed out that current AL cannot improve few-shot learning models significantly compared to random sampling. \cite{Pezeshkpour2020on} The main reason could be that AL and ProtoNet are processed separately in their work, i.e., AL is only used for preparing the support set of ProtoNet. Our proposed novel algorithm - Active Transfer Prototypical Network (ATPN), focuses on training a ProtoNet enhanced AL learner. The idea is similar to \cite{WoodwardF17}, which introduces an active learner combining meta-learning and reinforcement learning to choose whether to label the instance by the annotator. But instead of reinforcement learning, we formulate a framework to integrate ProtoNet properly in the Al loop.

\section{Background}\label{background}
The proposed model consists of two machine learning techniques, namely 1) \textbf{AL} to iteratively select representative support set for ProtoNet; 2) \textbf{ProtoNet} to provide a more robust initial model and a more efficient learning mechanism. The learning efficiency is expected to be significantly higher and robust regardless of the performance of the size of the initial model. In this section, relevant background is given in detail.
\subsection{Prototypical Network}
The novel approach is mainly based on ProtoNet, where an embedding space is learned by computing distances to prototype representations of each class [7]. 

It requires a small representative labeled dataset $S = \{(x_1, y_1), \dots, (x_N, y_N)\}$ called support set, where $N$ is the number of examples and $x_i\in R^D$ refers to the $D$-dimensional feature vector. Assume there are $K$ different classes, $S_k$ denotes the subset of support set with label $k$. Per-class prototypes $c_k$ are calculated by averaging the $S_k$ in $M$-dimensional embedding space using an encoder $f_\phi:R^D\rightarrow R^M$. The classification probabilities $p_{\phi}$ can be calculated based on a distance function $d:R^M\times R^M\rightarrow [0,+\infty)$, e.g., Euclidean distance is used in this paper. Since the original paper of ProtoNet \cite{snellPrototypicalNetworksFewshot2017} demonstrates that compared to cosine distance and other distance metrics, using Euclidean distance can significantly improve results. 
\begin{equation}
\label{eqn:classProb}
p_{\phi}(y=k|x) = \frac{\exp(-d(f_{\phi}(x), c_k))}{\sum_{i=1}^K\exp(-d(f_{\phi}(x), c_i))}, \ \ \ \text{where}\ \ \ \ \ \ c_k = \frac{1}{|S_k|}\sum_{(x_i, y_i)\in S_k}f_\phi(x_i)
\end{equation}
The learnable parameters $\phi$ can be then computed by minimizing the cost function $J(\phi)=-\log_{p_\phi}(y=k|x)$ on a training set $D$, where $|D|\gg |S|$.

\subsection{Query Strategy}\label{querystrategy}
The AL algorithms can query the data based on \textit{uncertainty sampling} [16], \textit{query-by-committee} [17], or \textit{expected model change} [18]. Among them, query-by-committee requires an ensemble model, and expected model change is computationally expensive. Therefore, uncertainty sampling is chosen due to its simplicity and interpretability.

\textbf{Least Confidence} The most straightforward measure is the uncertainty of classification  defined by $U(x) = 1-P(\hat{x}|x)$ \cite{lewis1994sequential}, where x is the data point to be classified and $\hat{x}$ is the prediction with the highest classification probability. It selects the instances that the model is least confident in prediction based on current estimated classification probability.

\textbf{Margin Sampling} Classification margin \cite{lewis1994sequential} is the difference between the probability of the first and second most likely prediction, which is defined by $M(x) = P(\hat{x_1}|x) - P(\hat{x_2}|x)$, where $\hat{x_1}$ and $\hat{x_2}$ are the classes with first and second-highest classification probability. It assumes that the most informative instances fall within the margin.

\textbf{Entropy Sampling} In information theory, the terminology "entropy" \cite{shannon1948} refers to the average level of information or uncertainty inherent to the variable's possible outcomes. Inspired by this, the classification uncertainty can also be represented by $H(x) = -\sum_{i=1}^{n}P(x_i)\log P(x_i)$, where $p_k$ is the probability of the sample belonging to the kth class, heuristically, the entropy is proportional to the average number of guesses one has to make to find the true class. The closer the distribution to uniform, the larger the entropy. It means that the instances with similar classification probabilities for all classes have a higher chance of being selected while querying.

\textbf{Batch-Mode Sampling} More instances can be queried in a single AL iteration; \textit{ranked batch-mode query} \cite{CARDOSO2017313} is used for batch-mode sampling. The score of an instance can be calculated by $score=\alpha (1-\Phi(x, X_{labeled}))+(1-\alpha)U(x)$, where $\alpha=\frac{|X_{unlabeled}|}{|X_{unlabeled}|+|X_{labeled}|}$, $X_{labeled}$ is the labeled dataset, $U(x)$ is the least confidence measurement, and $\Phi$ is cosine similarity in the implementation. The first term measures the diversity of the current instance concerning the labeled dataset. The uncertainty of predictions for $x$ is considered in the second term. After scoring, the highest scored instance is put at the top of a list. Then, the instance is removed from the pool. Finally, the score is recalculated until the desired amount of instances is selected.

\section{Methodology}\label{methodology}
In this section, we demonstrate the detailed structure design of ATPN and its variants for ablation study.
\subsection{Active Transfer Prototypical Network}\label{ATFN}
With a large number of unlabeled measurements, such as in the automotive industry use cases, the entire unlabeled dataset should be evaluated to find the most informative instances at each single AL iteration. Therefore, a pool-based AL learning strategy is more suitable than stream-based selective sampling. In a pool-based scenario, AL attempts to find a small set of representative data $L$ from a large unlabeled data pool $U$ such that it can achieve satisfactory performance with much less labeling effort, i.e., $|L| \ll |U|$. Algorithm \ref{alg:main} is designed to reach the minimal $|L|$. A pre-trained encoder $f_\phi$ is used instead of learning $\phi$ on $D$ to avoid updating the entire parameter set in every AL iteration. Only the parameters of the feature layer $\phi'$, i.e., the layer preceding the output layer of the encoder, is updated in AL iterations. $\hat{\phi}$ denotes the parameters frozen in AL iterations.
\begin{algorithm}[H]
\caption{\textit{Training Active Transfer Prototypical Network}}\label{alg:main}
\textbf{Input:} pre-trained encoder $f_{\phi}(x)$; unlabeled data pool $\mathbf{U}=\{x_1,\dots,x_m\}$
\begin{algorithmic}
    \State $\mathbf{L} \gets RandomSample(\mathbf{U}, 1)$ 
	\State $\mathbf{U} \gets \mathbf{U}\setminus\mathbf{L}$ \vspace{1mm}
	\State \textbf{Initialization of Feature Layer}
	\State $\mathbf{J_{\hat{\phi}}}\gets \sum_{(x_i, y_i)\in\mathbf{L}}(y_i - softmax(f_{\phi', \hat{\phi}}(x_i)))^2$
	\State $\hat{\phi}\gets argmin_{\hat{\phi}}\mathbf{J_{\hat{\phi}}}$\vspace{1mm}
	\State \textbf{Initialization of Prototypical Network}
	\State $c_k\gets\frac{1}{|\mathbf{L}_k|}\sum_{(x_i,y_i)\in\mathbf{L}_k}f_{\phi', \hat{\phi}}(x_i)$\vspace{1mm}
	\State \textbf{Active Learning Iterations}
	\For{$n$ in $\{1,\dots,N\}$}\vspace{1mm}
	\State $p_{\phi', \hat{\phi}}(y=k|\textbf{x}) \gets \frac{\exp(-d(f_{\phi', \hat{\phi}}(\textbf{x}), c_k))}{\sum_{i=1}^{K}\exp(-d(f_{\phi', \hat{\phi}}(\textbf{x}), c_{i}))}$ \vspace{1mm}
	\State $\mathbf{S} \gets QueryStrategy(\mathbf{U}, p_{\phi', \hat{\phi}})$\vspace{1mm}
	\State $\mathbf{L} \gets \mathbf{L}\bigcup\mathbf{S}$\vspace{1mm}
	\State $\hat{\phi}\gets argmin_{\hat{\phi}}\sum_{(x_i, y_i)\in\mathbf{L}}(y_i - softmax(f_{\phi', \hat{\phi}}(x_i)))^2$\vspace{1mm}
	\State $c_k\gets\frac{1}{|\mathbf{L}_k|}\sum_{(x_i,y_i)\in\mathbf{L}_k}f_{\phi', \hat{\phi}}(x_i)$ \vspace{1mm}
	\EndFor
\end{algorithmic}
\end{algorithm}

The ATPN starts with random initialization. One instance is randomly sampled from $U$ and labeled by an expert. The pseudo-label of the initial sampled point is predicted using the pre-trained encoder $f_{\phi}$. The probability is obtained by the softmax layer placed on top of the encoder. $\phi'$ is initially updated by minimizing the \textit{Mean Square Error} (MSE) of the initial point. $L$ is considered a support set for ProtoNet; therefore, the prototypes $c_k$ are initialized with $L$ through averaging $L_k$ in embedding space.

In the AL iterations, the classification probability $p_{\phi', \hat{\phi}}$ is estimated based on the negative Euclidean distance $d$ between the embedded input feature $f_{\phi', \hat{\phi}}$ and the prototypes $c_k$. Then, the query strategy evaluates the informativeness for every single instance in $U$ using $p_{\phi', \hat{\phi}}$ as input. Intuitively, the instances far away from the prototypes are regarded as informative instances. The most informative instance or set of instances $S$ is selected and added to $L$. 

Furthermore, the labeled dataset $L$ is used to update two parts of our algorithm. On the one hand, the prototypes are updated by averaging the embeddings in the new $L$ to increase the representativeness of each class. On the other hand, we fine-tune the encoder, i.e., update the parameters of the feature layer $\phi'$ by minimizing the MSE in $L$. It aims to preserve the learning ability of the algorithm even when the test data and the training data for the pre-trained encoder are not similar. Finally, the iteration process repeats $N$ times, which is the maximal query number. The maximal query number $N$ corresponds to the budget for data labeling.

\subsection{Active Prototypical Network Variants}
To investigate the influence of each relevant part in ATPN, we implement two other variants - \textit{Active Offline Prototypical Network} (OfflinePN) and \textit{Active Online Prototypical Network} (OnlinePN). The former does not update the encoder's feature layer $\phi'$ during AL iterations to observe the impact of fine-tuning; the latter removes the pre-trained encoder and trains the encoder entirely online with the support set $L$ to identify the improvement of the pre-trained encoder.

Algorithm \ref{alg:offline} removes the fine-tuning part from ATPN. In Algorithm \ref{alg:online}, there is no pre-trained encoder. The learnable parameters $\phi', \hat{\phi}$ are initialized randomly and updated online in AL iterations with support set $L$.
\begin{minipage}{0.49\textwidth}
\begin{algorithm}[H]
\caption{\textit{Training Active Offline Prototypical Network}}\label{alg:offline}
\textbf{Input:} pre-trained encoder $f_{\Phi}(x)$; unlabeled data pool $\mathbf{U}=\{x_1,\dots,x_m\}$ \vspace{2mm}
\begin{algorithmic}
    \State $\mathbf{L} \gets RandomSample(\mathbf{U}, 1)$ \vspace{1mm}
	\State $\mathbf{U} \gets \mathbf{U}\setminus\mathbf{L}$ \vspace{2mm}
	\State \textbf{Initialization of Prototypical Network}
	\State $c_k\gets\frac{1}{|\mathbf{L}_k|}\sum_{(x_i,y_i)\in\mathbf{L}_k}f_{\phi', \hat{\phi}}(x_i)$\vspace{2mm}
	\State \textbf{Active Learning Iterations}\vspace{1mm}
	\For{$n$ in $\{1,\dots,N\}$}\vspace{2mm}
	\State $p_{\phi', \hat{\phi}}(y=k|\textbf{x}) \gets \frac{\exp(-d(f_{\phi', \hat{\phi}}(\textbf{x}), c_k))}{\sum_{i=1}^{K}\exp(-d(f_{\phi', \hat{\phi}}(\textbf{x}), c_{i}))}$ \vspace{2mm}
	\State $\mathbf{S} \gets QueryStrategy(\mathbf{U}, p_{\phi', \hat{\phi}})$\vspace{2mm}
	\State $\mathbf{L} \gets \mathbf{L}\bigcup\mathbf{S}$\vspace{2mm}
	\State $c_k\gets\frac{1}{|\mathbf{L}_k|}\sum_{(x_i,y_i)\in\mathbf{L}_k}f_{\phi', \hat{\phi}}(x_i)$ \vspace{2mm}
	\EndFor
\end{algorithmic}
\end{algorithm}
\end{minipage}
\hfill
\begin{minipage}{0.49\textwidth}
\begin{algorithm}[H]
\caption{\textit{Training Active Online Prototypical Network}}\label{alg:online}
\textbf{Input:} unlabeled data pool $\mathbf{U}=\{x_1,\dots,x_m\}$
\begin{algorithmic}
    \State $\mathbf{L} \gets RandomSample(\mathbf{U}, 1)$ 
	\State $\mathbf{U} \gets \mathbf{U}\setminus\mathbf{L}$ \vspace{1mm}
	\State \textbf{Initialization of Encoder}
    \State $\phi', \hat{\phi}\gets Randomly\ initialize\ f_{\phi', \hat{\phi}}(x_i)$\vspace{1mm}
	\State \textbf{Initialization of Prototypical Network}
	\State $c_k\gets\frac{1}{|\mathbf{L}_k|}\sum_{(x_i,y_i)\in\mathbf{L}_k}f_{\phi', \hat{\phi}}(x_i)$\vspace{1mm}
	\State \textbf{Active Learning Iterations}
	\For{$n$ in $\{1,\dots,N\}$}\vspace{1mm}
	\State $p_{\phi', \hat{\phi}}(y=k|\textbf{x}) \gets \frac{\exp(-d(f_{\phi', \hat{\phi}}(\textbf{x}), c_k))}{\sum_{i=1}^{K}\exp(-d(f_{\phi', \hat{\phi}}(\textbf{x}), c_{i}))}$ \vspace{1mm}
	\State $\mathbf{S} \gets QueryStrategy(\mathbf{U}, p_{\phi', \hat{\phi}})$\vspace{1mm}
	\State $\mathbf{L} \gets \mathbf{L}\bigcup\mathbf{S}$\vspace{1mm}
	\State $\phi', \hat{\phi}\gets argmin_{\phi', \hat{\phi}}\{-\log(p_{\phi', \hat{\phi}}(y=k|\textbf{x}))\}$\vspace{1mm}
	\State $c_k\gets\frac{1}{|\mathbf{L}_k|}\sum_{(x_i,y_i)\in\mathbf{L}_k}f_{\phi', \hat{\phi}}(x_i)$ \vspace{1mm}
	\EndFor
\end{algorithmic}
\end{algorithm}
\end{minipage}

\section{Experimental Settings}\label{exp}
The algorithms are validated on the \textit{UCI HAR} \cite{HAR} \textit{\& HAPT} dataset \cite{HAPT} and a Bosch braking maneuver dataset. We run the algorithms five times and record the accuracy of each AL iteration. Then, the average accuracy is calculated, and the confidence interval is determined using \textit{bootstrapping} method \cite{diciccio1996bootstrap}. Finally, we plot an accuracy curve in terms of query number to evaluate the model performance.
\subsection{UCI HAR\&HAPT Dataset}
HAR and HAPT datasets are human activity recognition databases built from the recordings of subjects performing activities of daily living while carrying a waist-mounted smartphone with embedded inertial sensors. The trials of both datasets were conducted on 30 participants ranging in age from 19 to 48 years old, who wore a smartphone (Samsung Galaxy S II) around their waist during the experiment. With the device's built-in accelerometer and gyroscope, 3-axial linear acceleration and 3-axial angular velocity were recorded at a rate of 50Hz. The HAR dataset contains six basic human activities: three static postures - \textit{standing}, \textit{sitting}, and \textit{lying} and three dynamic activities - \textit{walking}, \textit{walking downstairs}, and \textit{walking upstairs}. While HAPT dataset further includes postural transitions that occurred between the static postures. These are \textit{stand-to-sit, sit-to-stand, sit-to-lie, lie-to-sit, stand-to-lie, and lie-to-stand}. The sensor signals (accelerometer and gyroscope) are pre-processed by applying median filters, and a third-order low pass Butterworth filter with a corner frequency of 20 Hz to remove noise. Then it is sampled in fixed-width sliding windows of 2.56 sec and 50\% overlap (128 readings/window) to standardize the signal length.

\subsection{BOSCH Braking Maneuver Dataset}
To validate the investigated algorithms, they were tested on a real-world industrial database - BOSCH \textit{Automated Maneuver Detection} (AMD) database. It contains braking maneuvers for software functions, e.g., \textit{Anti-lock Braking System} (ABS) and \textit{Electronic Brake Force Distribution} (EBD). This database aims to train a general feature extractor to automate the maneuver detection task. The measurements were initially collected from the quality center database, the official data pool for storing brake maneuver measurements from customers and platform development projects. These measurements were mainly measured by sensors in test vehicles. 10 ABS maneuver types and 5 EBD maneuver types were used in this paper. Each maneuver type contains hundreds of measurements. There are in total 12277 measurements, 8536 ABS measurements, and 3741 EBD measurements. Concretely, each measurement records braking events under different scenarios such as different road friction coefficients and pavement Slope. 18 relevant sensor signals, including braking pressure, wheel velocity, etc., are selected as input for our model. All signals are re-sampled to length 4000 using linear interpolation to get a uniform input dimension.

\subsection{Pre-trained Encoder}
We use \textit{1D Convolutional Neural Network} (CNN) \cite{LeCun1998} as a time-series encoder; theoretically, it can be replaced by any neural network capable of extracting solid features from time-series such as \textit{Long short-term Memory} (LSTM) \cite{HochSchm97}. The objective of a pre-trained encoder in our model is to provide a solid initial model for the AL loop. As long as the pre-trained encoders extract common features, what architecture is used and the hyperparameter-tuning has a minor effect on the final result. Since the feature layer will be fine-tuned in the AL loop, the adaption from different pre-trained encoders is efficient. Additionally, this paper focuses on investigating the overall combination concept of Al and ProtoNet. Therefore we select an efficient hyperparameter setting to reduce the experiment time.

For the UCI HAR dataset, the encoder contains two 1D convolutional layers with 32 filters, 7 kernels, and \textit{Rectified Linear Unit} (ReLU) \cite{relu} as activation function, followed by a Maxpooling layer with pool size 2 and 0.5 rate dropout layer. The encoder ends up with a flatten layer and a softmax layer. We use Adam as an optimizer, Categorical Cross Entropy as a loss function, and run 10 epochs with batch size 32.

Further hyperparameter tuning and more epochs lead to an improvement for the pre-trained encoder. Still, our experiments show that the difference in overall performance with respect to the hyperparameter tuning is minor. However, the training data used for the encoder is relevant (see section \ref{pre-train exp}).

\subsection{Experiments}\label{expsetting}
We conduct four experiments on the UCI HAR\&HAPT dataset to investigate the algorithm performance in terms of algorithm structure, query strategy, and pre-trained encoder. The first experiment compares all algorithm variants to standard AL and passive learning. 1D CNN is used as a machine learning model in the AL framework with the least confidence sampling as a query strategy. The same AL architecture with random sampling as query strategy refers to passive learning in this context. The second and third experiment attempts to investigate the impact of query strategy. The query strategies in section \ref{querystrategy} are applied in the experiment. In the last experiment, the pre-trained encoder is trained on datasets with different similarities to the test dataset. For "weak encoder," the encoder is trained on a different dataset, namely the bosch braking maneuver dataset. "Normal encoder" uses UCI HAR as training data, and "strong encoder" further includes some UCI HAPT data.

\begin{figure}[ht]
\centerline{\includegraphics[width=0.49\textwidth]{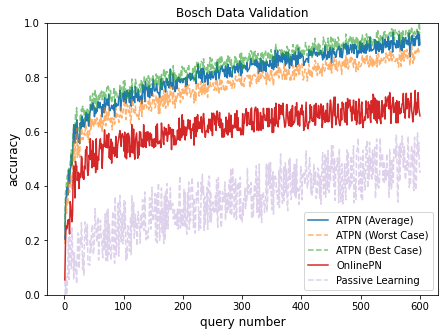}}
\caption{The learning curves of ATPN, OnlinePN and passive learning for the Bosch brake maneuver dataset. For the best algorithm - ATPN, experiments with different pre-trained encoders and test sets are conducted; the worst \& best case and the average learning curve are shown in the figure.}
\label{fig:bosch}
\end{figure}

To identify the improvement of the novel algorithm in a real-world application, only ATPN and OnlinePN are compared to passive learning, which is commonly used in industry. The best algorithm settings based on the previous experiment are used for the Bosch dataset. The robustness of the pre-trained encoder is vital for the model. Therefore, the braking measurements of 9 braking maneuvers are randomly selected from 15 maneuvers as training data for the encoder, and the measurements of the rest 6 maneuvers are considered test data. This random selection repeats five times. The best and worst case is recorded in Fig. \ref{fig:bosch}.

\section{Results}
In this section, To better describe the observations, two main processes can be defined in each learning curve of the algorithm. 
\begin{itemize}
    \item \textbf{Exploration Process}: during this process, AL attends to learn a reliable model to estimate classification probability for the query strategy.
    \item \textbf{Exploitation Process}: AL has learned a sufficient model to predict reliable informativeness of unlabeled data. More labeled instances do not change the unlabeled instances’ informativeness significantly. Therefore, the accuracy will increase slowly in this process. The slope represents the learning efficiency of AL.
\end{itemize}

\begin{figure}[ht]
\centerline{\includegraphics[width=1.02\textwidth]{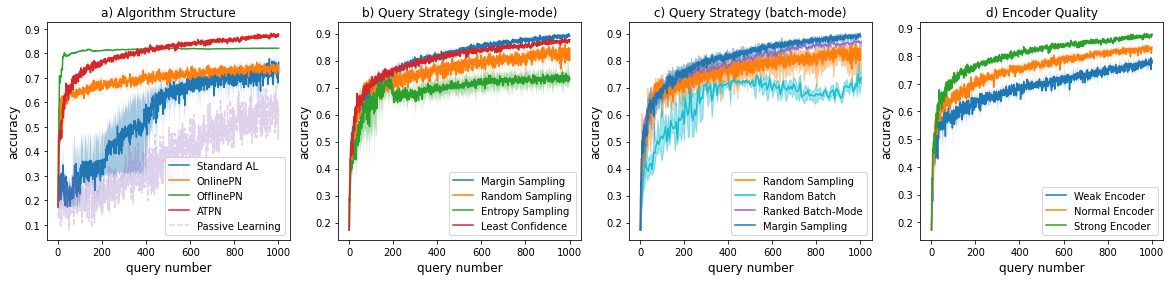}}
\caption{The results of experiments undertaken on the UCI HAR\&HAPT dataset. a) different versions of active ProtoNet are compared with standard AL and passive learning. b) different single-mode query strategies are tested for ATPN. c) batch-mode query strategies are compared with margin sampling and random sampling. d) pre-trained encoders with different generalization levels are tested for ATPN with margin sampling.}
\label{fig:HAR}
\end{figure}


\subsection{Algorithm Structure}
Fig. \ref{fig:HAR} a. shows that the learning curve of passive learning has an overall linear growth trend with substantial deviation. Standard AL improves learning efficiency by requiring fewer queries to achieve the same classification accuracy. However, standard AL has a noisy exploration process. The results of all AL-ProtoNet variants follow a pattern where the curve has a significantly larger slope in the exploration process than in the exploitation process. OnlinePN and ATPN both reach 60\% accuracy within 50 queries. OnlinePN has a flatter exploitation curve and converges with standard AL after 1000 queries. In contrast, the accuracy of ATPN keeps increasing visibly in the exploitation process. Surprisingly, the exploration process of OfflinePN is most efficient, but the accuracy improvement stops at 80\% within very few queries.

The comparison between OnlinePN and standard AL indicates the structure of ProtoNet provides a more efficient exploration process; they differ only in the machine learning algorithm, i.e., OnlinePN uses ProtoNet instead of 1D CNN as an AL algorithm. OfflinePN leverages the prior knowledge in a pre-trained encoder and further optimizes the exploration efficiency, but the training dataset's quality strictly limits the optimal accuracy. Fine-tuning during the AL iterations balances the exploration and exploitation. Therefore, ATPN outperforms other variants in terms of the overall learning ability.

\subsection{Query Strategy}
The influence of different query strategies is investigated in Fig.\ref{fig:HAR} b. and c. for ATPN. Different query strategies does not change the shape of the learning curve significantly. Margin sampling and least confidence sampling perform slightly better than random sampling and have comparable behavior in the exploration process. Margin sampling has marginally higher learning efficiency in the exploitation process. In contrast, entropy sampling provides worse learning performance in the exploration and exploitation process than random sampling. There is a unusual plunge close to the critical point between exploration and exploitation. Furthermore, a batch-mode query strategy could increase the learning efficiency because it considers both informativeness and diversity simultaneously. However, Figure \ref{fig:HAR} c) does not completely support this hypothesis. Random batch sampling performs worse than random sampling because an unsuitable batch could cause a more considerable bias to the model, leading to a more fluctuating learning curve. Therefore, according to these experiments, margin sampling is best suited for ATPN.

\subsection{Pre-trained Encoder}\label{pre-train exp}
Fig. \ref{fig:HAR} d) demonstrates that the change in the exploration and exploitation process due to different pre-trained encoders is minor. The learning curves seem to be only shifted vertically for different encoders. In other words, the pre-trained encoder influences the turning point from the exploration to the exploitation process.

\subsection{Validation on Bosch dataset}
To validate the novel algorithm in the real-world dataset, we mainly focus on investigating ATPN. The goal is to indicate the improvement of the novel algorithm compared to the algorithm currently used in the automotive industry, i.e., passive learning. The validation of ATPN is done with different test settings as described in section \ref{expsetting}. The average learning curve shows similar behavior as for the UCI HAR dataset, which verifies the generalization performance of the novel algorithm to a more complex time-series dataset. We visualize the worst case and best case for the ATPN learning curve, demonstrating the algorithm's robustness in practice. Even though it has been shown that a pre-trained encoder has an impact on the overall learning performance in previous results, the deviation due to the pre-trained encoder is acceptable in practice. ATPN reaches over 90\% accuracy with less than 10\% of labeling effort on average.

Moreover, in an extreme case, the training data for the pre-trained encoder and the test set can be independent. ATPN and OnlinePN are equivalent in this circumstance since the encoder is not pre-trained regarding the current task. The learning curve of OnlinePN indicates the lowest bound of the learning performance. The results show that ATPN significantly improves even if the pre-trained encoder is not well trained.

\section{Conclusions and Future Work}\label{conclusion}
We propose a novel AL framework incorporating a fine-tuning ProtoNet for time-series classification. The investigation of different combinations has shown that ATPN has the most comprehensive learning ability, including learning efficiency and robustness. Furthermore, the research has shown that the quality of the pre-trained encoder can influence the learning performance of ATPN. However, the encoder can keep the deviation of the learning curve within acceptable limits in practice. Moreover, margin sampling provided the best learning performance in both datasets. This paper indicates that ATPN can address the trade-off dilemma between the training data size for the initial model and the query number. It also provides a more efficient and robust learning process for a pool-based AL strategy. 

A training set of similar tasks is required to obtain the optimal learning curve using ATPN. Developing a similarity metric for the labeled training data and the current data could address this problem. Combining an unsupervised encoder such as Autoencoder would also be fruitful for further work. More advanced AL strategies, e.g., query by committee, could improve the learning performance. Notwithstanding these limitations, the study indicates that the proposed algorithm can reduce the effort for labeling time-series data dramatically in practice.

\bibliography{main}
\bibliographystyle{elsarticle-harv}

\end{document}